\documentclass{article} 

\usepackage{capt-of}
\usepackage{algorithm}
\usepackage{algpseudocode}
\usepackage{booktabs}       
\usepackage{wrapfig}
\usepackage{amsfonts}       
\usepackage{nicefrac}       
\usepackage{microtype}      
\usepackage{xcolor}         
\usepackage{graphicx}
\usepackage{amsmath}

\newtheorem{theorem}{Theorem}[section]

\newtheorem{lemma}[theorem]{Lemma}

\usepackage{iclr2024_conference,times}

\usepackage{amsmath,amsfonts,bm}

\def\eqref#1{equation~\ref{#1}}

\def\1{\bm{1}}

\DeclareMathAlphabet{\mathsfit}{\encodingdefault}{\sfdefault}{m}{sl}
\SetMathAlphabet{\mathsfit}{bold}{\encodingdefault}{\sfdefault}{bx}{n}

\usepackage{hyperref}
\usepackage{url}

\title{Understanding Contrastive Learning Through the Lens of Margins}
\iclrfinalcopy

\author{%
  Daniel Rho \\
  KT \\
  \And
  TaeSoo Kim \\
  KT, Sungkyunkwan University \\
  \AND
  Sooill Park \\
  Hyundai Mobis \\
  \And
  Jaehyun Park \\
  NC Soft \\
  \And
  JaeHan Park \\
  KT
}

\begin{document}

\maketitle

\begin{abstract}
Contrastive learning, along with its variations, has been a highly effective self-supervised learning method across diverse domains.
Contrastive learning measures the distance between representations using cosine similarity and uses cross-entropy for representation learning.
Within the same framework of cosine-similarity-based representation learning, margins have played a significant role in enhancing face and speaker recognition tasks.
Interestingly, despite the shared reliance on the same similarity metrics and objective functions, contrastive learning has not actively adopted margins.
Furthermore, decision-boundary-based explanations are the only ones that have been used to explain the effect of margins in contrastive learning.
In this work, we propose a new perspective to understand the role of margins based on gradient analysis.
Based on the new perspective, we analyze how margins affect gradients of contrastive learning and separate the effect into more elemental levels.
We separately analyze each and provide possible directions for improving contrastive learning.
Our experimental results demonstrate that emphasizing positive samples and scaling gradients depending on positive sample angles and logits are the keys to improving the generalization performance of contrastive learning in both seen and unseen datasets, and other factors can only marginally improve performance.


\end{abstract}

\section{Introduction}
\label{sec:intro}

Self-supervised learning (SSL), or unsupervised learning, has attracted a lot of attention, succeeding in a range of fields~\cite{wu2018unsupervised,infoNCE(CPC),devlin2018bert,dosovitskiy2020image,NEURIPS2021_27debb43}.
Contrastive learning~\cite{wu2018unsupervised,infoNCE(CPC)} is one of the universal SSL frameworks~\cite{infoNCE(CPC)} not relying on domain-specific assumptions.
It learns instance-level relationships between samples using the similarity function (cosine similarity) and cross-entropy.
Pretraining frameworks based on contrastive learning have been applied in various fields, ranging from the image ~\cite{MoCo,SimCLR,SwAV,BYOL} to the audio domain~\cite{infoNCE(CPC),hsu2021hubert}.

As contrastive learning does, face and speaker recognition tasks share the same cosine similarity and softmax loss-based losses~\cite{AdaCos,CurricularFace,MagFace,AdaFace,ElasticFace,ECAPA-TDNN}.
To improve identification accuracy, these domains usually exploit margins, which are known to increase inter-class distance and decrease intra-class distance in face and speaker recognition tasks.
To be more specific, they add margins to the decision boundary by adding margins to the angles or the logits of positive samples (Eq.~\ref{eq:delta}).
Based on the commonality, there have been recent attempts to benefit from the characteristic of margins in contrastive learning~\cite{Zhan2022,zhang2022incorporating}.
These studies demonstrate that margins can improve contrastive learning-based tasks.
Notwithstanding the difference between face recognition and contrastive learning, the explanation of how margins work remains solely based on the explanation from the face recognition domain.

Thus, our work starts with the following question: how do margins affect contrastive learning?
It remains unclear how margins affect contrastive learning in the absence of class labels, which are commonly used in face recognition tasks to explain the effect of margins.
Thus, our work aims to investigate how margins affect contrastive learning-based representation learning through gradient analysis without relying on classification- or decision-boundary-based explanations.
We use the generalized contrastive learning loss~\cite{infoNCE(CPC)} to incorporate margins (Sec.~\ref{ssec:generalized-margin-loss}).
Then, we analyze the gradient of the loss to identify the effect of margins on gradients (Sec.~\ref{sec:gradient-analysis}).

Through gradient analysis, we found that margins affect representation learning in four ways.
First, it emphasizes positive samples.
Second, margins reduce the gradients of distant positive samples.
Third, it scales gradients by the ratio of sums of exponentiated logits without and with margins, which is affected by both the angle and logits of positive samples.
Lastly, margins alleviate the slowdown effect of gradients when the estimated probability approaches the target probability.
Based on the analysis, we separately explored each effect.
We experimentally showed that emphasizing positive samples and scaling gradients by the ratio of sums of exponentiated logits are important for improving contrastive learning.

Contrastive learning is based on identity discrimination~\cite{wu2018unsupervised}, relying on the idea that representations of the same identity should cluster together while those from different identities should separate.
Aside from this core idea, there are a lot of potential directions for improvement.
We believe that understanding how margins affect contrastive learning can provide direction for future improvement in contrastive learning and SSL, not only helping us exploit margins in contrastive learning.
Not only that, we believe our new perspective on margins could help better understand the role of margins even in other tasks, including face recognition.

To summarize, our contributions are threefolds:

$\bullet$ We provide a new perspective on the role of margins in cosine similarity-based representation learning through gradient analysis.

$\bullet$ We show that margins induce a mixture of effects, separate each effect, and experimentally validate the efficacy of each separated effect and provide its limitations.

$\bullet$ Our experimental results demonstrate that emphasizing positive samples and scaling gradients based on positive sample angles and logits are the keys to improving the generalization performance of contrastive learning in both seen and unseen datasets. 


\section{Related Works}
\label{sec:related_works}

\subsection{Contrastive Learning}
\label{ssec:self-supervised-learning}
Contrastive learning (InfoNCE)~\cite{infoNCE(CPC)} aims to learn instance-level relationships between samples using a similarity function (cosine similarity) and cross-entropy.
Contrastive learning enforces neural networks to generate close representations for positive samples (different views of the same image) and distant representations for negative samples (views from different images)~\cite{wu2018unsupervised}.
Several InfoNCE-based SSL methods~\cite{SimCLR,SwAV,BYOL,MoCo,SimSIAM} have been proposed.
For example, \textit{MoCo} \cite{MoCo,MoCov2,MoCov3} proposed using a teacher model for generating different latent representations of the same samples.
\textit{SimCLR}~\cite{SimCLR,SimCLRv2} demonstrated that augmentations play a critical role in contrastive learning frameworks.
\textit{BYOL}~\cite{BYOL}, another variety of InfoNCE, proposed using only positive samples.

Several works have analyzed the properties and limitations of contrastive learning~\cite{Wang_2021_CVPR,zhang2022incorporating,Wang2022,debiased}.
\citet{Wang_2021_CVPR} analyzed the role of temperature $\tau$ in contrastive learning, and found that contrastive learning focuses on nearby samples.
Similarly, \citet{zhang2022incorporating} argue that contrastive learning is robust to long-tail distribution.
There are additional studies to uncover the weak spots of contrastive learning and suggest remedies.
\citet{Wang2022}, for example, showed that contrastive learning tends to ignore non-shared information between views, resulting in performance degradation in some downstream task.
Meanwhile, \citet{debiased} demonstrate that contrastive learning is biased and thus de-biasing contrastive learning loss can improve representation learning.

\subsection{Margin softmax loss}
\label{ssec:margin_softmax}
Face and speaker recognition tasks determine whether two given samples are representations of the same identity by comparing extracted feature vectors.
These tasks mainly use the same cosine-similarity-based softmax loss as contrastive learning, the difference being using known identity information as class labels during training.
Margins have been a successful method for widening inter-class, or inter-identity, distance and narrowing down intra-class distance~\cite{CosFace,ArcFace}.
To improve recognition performance, several variety of margins have been proposed, including adaptive~\cite{AdaCos,CurricularFace,MagFace,AdaFace} and stochastic margins~\cite{ElasticFace}.

There have been attempts to introduce margins also in contrastive learning (or InfoNCE)~\cite{Zhan2022,zhang2022incorporating}.
\citet{Zhan2022} showed that margins can enhance feature discriminability, and \citet{zhang2022incorporating} showed that margins could be used to reduce population bias.
Yet, no work has delved into gradient levels to analyze how and why margins work, and the explanations on the role of margins remain in the feature-discriminative aspects of margins.
In this work, we aim to understand the margin through gradient analysis and identify the complex effects that margins have on contrastive learning.

\section{Generalized Margins for Contrastive Loss}
\label{sec:method}
In this section, we generalize InfoNCE~\cite{infoNCE(CPC)} loss and include margins in order to analyze its effect on gradients.
For consistency, we will use notations $j$ and $k$ for arbitrary indices, and notations $l$ and $h$ for positive and negative sample indices, respectively.

\subsection{Generalized InfoNCE} 
\label{ssec:infonce-loss}

The InfoNCE loss function can be represented as follows:
\begin{equation}
    \tilde{q}_{ij} = \frac{\exp(sim(z_i, z_j)/\tau)}{\sum_{k \in \mathcal{X}}{\exp(sim(z_i, z_k)/\tau)}}, \quad \tilde{\mathcal{L}}_i = -\sum_{j \in \mathcal{X}}{p_{ij} \log \tilde{q}_{ij}}.
    \label{eq:infonce-loss}
\end{equation}
$z_i$ denotes a latent feature of input $i$. 
$\mathcal{X}$ denotes the set of samples in a mini-batch, and $p_{ij}$ denotes the target probability of two identities (i and j) being equal.
$sim(\cdot, \cdot)$ is a similarity function.
Given that cosine similarity is used as a similarity function, $\tilde{q}_{ij}$  can be rewritten as follows:
\begin{equation}
    \theta_{ij} = \arccos(sim(z_i, z_j)),\quad \tilde{\delta}_{ij} = \cos(\theta_{ij}) / \tau, \quad \tilde{q}_{ij} = \exp(\tilde{\delta}_{ij}) / \sum_{k \in \mathcal{X}}{\exp(\tilde{\delta}_{ik})}
    \label{eq:delta-ori}
\end{equation}
$\theta_{ij}$ denotes the angle between two normalized latent features.

Other contrastive learning techniques, like BYOL~\cite{BYOL}, that exclusively use positive samples cannot be covered by Eq.~\ref{eq:infonce-loss}.
As a result, we generalize the equation by introducing $\beta$ to the denominator of $\tilde{q}_{ij}$ as proposed in BYOL~\cite{BYOL}.
By rewriting the equation, we get the following equation:
\begin{align}
    \tilde{\mathcal{L}}_i = -\sum_{j \in \mathcal{X}}{p_{ij} \log \frac{\exp(\tilde{\delta}_{ij})}{\beta\sum_{k \in \mathcal{X}}{\exp(\tilde{\delta}_{ik})}}}
    = -\sum_{j \in \mathcal{X}}{p_{ij}\tilde{\delta}_{ij}} + \beta \sum_{j \in \mathcal{X}}{p_{ij}\log \sum_{k \in \mathcal{X}}{\exp(\tilde{\delta}_{ik})}}.
\label{eq:generalized-eq}
\end{align}
If $\beta$ is non-zero, the loss uses both positive and negative samples.
Otherwise, only positive samples will be used for training.
Therefore, this equation can generalize any contrastive learning-based SSL method, such as MoCo, SimCLR, and BYOL.

\subsection{Including margins in InfoNCE}
\label{ssec:generalized-margin-loss}
There are two types of margins; angular margin $m_1$ and subtractive margin $m_2$.
Angular margin $m_1$ is added to the angle between two representations $\theta_{ij}$, and subtractive margin $m_2$ is subtracted to the logits.
These margins are added only to positive samples.
After including margins, we can rewrite the logits and the estimated probability as follows:
\begin{equation}
    \delta_{ij} = (\cos(\theta_{ij} + m_1p_{ij}) - m_2p_{ij})/\tau,\quad q_{ij} = exp(\delta_{ij}) / \sum_{k \in \mathcal{X}}{exp(\delta_{ik})}
    \label{eq:delta}
\end{equation}
Likewise, Eq.~\ref{eq:generalized-eq} can be rewritten as follows:
\begin{equation}
    \mathcal{L}_i = -\sum_{j \in \mathcal{X}}{p_{ij}\delta_{ij}} + \beta \sum_{j \in \mathcal{X}}{p_{ij}\log \sum_{k \in \mathcal{X}}{\exp(\delta_{ik})}}.
    \label{eq:generalized-eq-with-margin}
\end{equation}
Eq.~\ref{eq:generalized-eq-with-margin} can generalize to various SSL methods, including MoCo, SimCLR and BYOL by simply setting margins to zeros.

\begin{figure}[]
\centering
\includegraphics[width=0.95\linewidth]{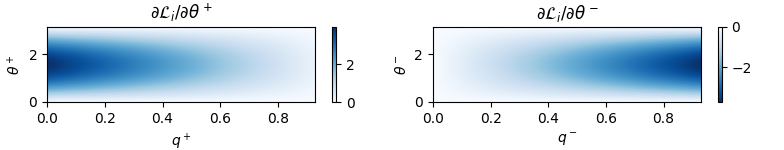}
\vspace{-0.5cm}
\caption{Gradient magnitudes of contrastive learning loss without margins (Eq.~\ref{eq:generalized-eq}). $\beta$, and $\tau$ was set to 1, and 0.25, respectively.
$q^+$ and $q^-$ denote the estimated probability of positive and negative samples. $\theta^+$ and $\theta^-$ refer to the angles of positive and negative samples.}
\label{fig:org-grad}
\end{figure}

\begin{figure*}[]
\centering
\includegraphics[width=0.95\textwidth]{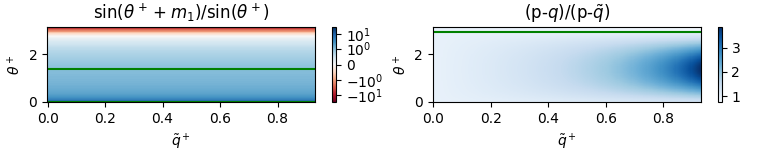}
\vspace{-0.4cm}
\caption{The gradient multipliers pertaining to the angular margin $m_1$ (Eq.~\ref{eq:scales}).
$\beta$, $\tau$, and $m_1$ was set to 1, 0.25, and 0.4, respectively.
The left figure shows the map that applies only to positive samples, while the right figure illustrates the multiplier map that applies to both positive and negative samples.
The green lines indicate where the weight is 1. Best viewed in color.
}
\label{fig:m1-scale}
\end{figure*}

\section{Gradient Analysis}
\label{sec:gradient-analysis}

As Fig.~\ref{fig:org-grad} illustrates, the magnitude of the gradient without margins $\partial \tilde{\mathcal{L}}_i/\partial \theta_{ij}$ diminishes as the estimated probabilities $q_{ij}$ of both positive and negative samples approach their target probabilities $p_{ij}$. 
To examine how margins affect representation learning, we compare the derivative of Eqs.~\ref{eq:generalized-eq} and \ref{eq:generalized-eq-with-margin} with respect to the angle $\theta_{ij}$.
We provide proofs of the following theorems and lemmas in the appendix.

\begin{theorem}
    The margins ($m_1,m_2$) scale the gradient with respect to an angle $\theta_{ij}$ by $\sin(\theta_{ij} + m_1p_{ij})/\sin(\theta_{ij}) \cdot  (p_{ij} - \beta q_{ij})/(p_{ij} - \beta \tilde{q}_{ij})$. That is,
\begin{equation}
    \frac{\partial \mathcal{L}_i}{\partial \theta_{ij}} = \frac{\partial \tilde{\mathcal{L}}_i}{\partial \theta_{ij}} \cdot \frac{\sin(\theta_{ij} + m_1p_{ij})}{\sin(\theta_{ij})} \cdot \frac{p_{ij} - \beta q_{ij}}{p_{ij} - \beta \tilde{q}_{ij}}.
\label{eq:scales}
\end{equation}
\label{theorem:main}
\end{theorem}
This shows that margins scale gradients through two terms; $\sin(\theta_{ij} + m_1p_{ij})/ \sin(\theta_{ij})$ and $(p_{ij} - \beta q_{ij})/ (p_{ij} - \beta \tilde{q}_{ij})$.

\begin{lemma}
    If $\beta$ equals one and $p_{ij}$ is either zero or one, $(p_{ij} - \beta q_{ij})/ (p_{ij} - \beta \tilde{q}_{ij}) = \sum \exp(\tilde{\delta}_{ij}) / \sum \exp(\delta_{ij}) = (\exp(\tilde{\delta}_{il})/q_{il}) \ /\ (\exp(\delta_{il}) + \exp(\tilde{\delta}_{il})(1/q_{il}-1))$.
\label{lemma:ratios}
\end{lemma}
That is, it is equal to the ratio of the sums of exponentiated logits without and with margins.
Based on these, we will analyze how angular margin $m_1$ and subtractive margin $m_2$ affect gradients in the following subsections.

\subsection{Angular margin}
\label{sssec:ga-angular-margin}

As Theorem~\ref{theorem:main} implies, the angular margin $m_1$ multiplies the gradients by two terms: $\sin(\theta_{ij} + m_1p_{ij})/ \sin(\theta_{ij})$ and  $(p_{ij} - \beta q_{ij})/ (p_{ij} - \beta \tilde{q}_{ij})$.
The first term $(p_{ij} - \beta q_{ij})/ (p_{ij} - \beta \tilde{q}_{ij})$ emphasizes positive samples and it also de-emphasizes positive samples as the angle $\theta_{il}$ increases.
The second term $\sin(\theta_{ij} + m_1p_{ij})/ \sin(\theta_{ij})$ scales gradients of both positive and negative samples.
Fig.~\ref{fig:m1-scale} visualizes these two terms.
We will elaborate on them in the following sections and experimentally verify the efficacy of each component in later sections.

As shown in Fig.~\ref{fig:m1-scale}, $\sin(\theta_{ij} + m_1p_{ij})/ \sin(\theta_{ij})$ drastically increases the scale of positive sample gradients.
Moreover, the gradient of a positive sample significantly diminishes as the angle $\theta_{il}$ widens. 
In short, the angular margin $m_1$ forces neural networks to focus on positive samples with small angles.

The second term $(p_{ij} - \beta q_{ij})/ (p_{ij} - \beta \tilde{q}_{ij})$ multiplies gradients of both positive and negative samples and has several characteristics.
First of all, it only relies on the angle ($\theta_{il}$ or $\theta^+)$ and the estimated probability ($\tilde{q}_{il}$ or $\tilde{q}^+$) of a positive sample only.
In addition, the weight increases as $\theta^+$ approaches $(\pi-m)/2$ and $\tilde{q}^+$ approaches one ($p_{il}$ or $p^+$).
Unfortunately, this term cannot be expressed as a product of two arbitrary functions, $f(\theta^+)$ and $g(\tilde{q}^+)$.
Therefore, we will use this ratio as it is.

\subsection{Subtractive margin}
\label{ssec:ga-subtractive-margin}

The subtractive margin $m_2$ directly suppresses the logits of positive samples $\delta_{il}$.
This attenuates the diminishing gradients as the estimated probability of a positive sample $\tilde{q}_il$ approaches one. 

\begin{lemma}
    If $\beta$ is 1 and $p_{ij}$ is either zero or one,  $\partial \mathcal{L}_i/\partial \theta_{ij}$ can be expressed as follows:
    \begin{align}
         \frac{\partial \mathcal{L}_i}{\partial \theta_{ij}} = \frac{\partial \tilde{\mathcal{L}}_i}{\partial \theta_{ij}}
        \frac{\sin(\theta_{ij} + m_1p_{ij})}{\sin(\theta_{ij})}
        \frac{1}{1 - (1-\exp((\cos(\theta_{il} + m_1) - \cos(\theta_{il})-m_2)/\tau))\tilde{q}_{il}},
        \label{eq:gradient-analysis-m2}
    \end{align}
where $\tilde{q}_{il}$ denotes the estimated probability of a positive sample $l$ without margins.
\label{lemma:subtractive}
\end{lemma}
Given that $1/(1 - (1-\exp((\cos(\theta_{il} + m_1) - \cos(\theta_{il})-m_2)/\tau))\tilde{q}_{il})$ increases as $\tilde{q}_{il}$ increases, $m_2$ gives more weight as $\tilde{q}_{il}$ approaches $p_{il}$, which is one.

\begin{lemma}
    As $m_2$ approaches infinity and $\beta$ is 1, $\lim_{m_2\to\infty} \partial \mathcal{L}_i/\partial \theta_{il} = \sin(\theta_{il} + m_1)/\tau$.
\label{lemma:limit}
\end{lemma}
That is, the subtractive margin $m_2$ will make positive sample gradients independent of the estimated probabilities $\tilde{q}_{il}$ and thus of other negative samples.
Since the gradient multiplier also affects negative samples, we analyze this effect in Sec.~\ref{ssec:results-attenuation} in two ways: positive-sample-only and positive-sample-bound attenuation.

\section{Experimental Results}
\label{sec:exp}
In this section, we evaluate the impact of four effects: emphasizing positive samples (Sec.~\ref{ssec:results-emphasizing-positive}), weighting positive samples differently depending on their angles (Sec.~\ref{ssec:results-curvature}), scaling gradients by the ratio (Sec.~\ref{ssec:ratio}), and attenuating the diminishing gradient effect as the estimated probability of positive sample $q_{il}$ approaches the target probability $p_{il}$ (Sec.~\ref{ssec:results-attenuation}).
To this end, we conducted separate experiments on MoCov3~\cite{MoCov3}, SimCLR~\cite{SimCLR}, and BYOL~\cite{BYOL}. 
We also compared with the baselines (MoCo, SimCLR, and BYOL) using linear probing (only tuning a linear layer on top of the frozen pretrained backbone model) and transfer learning (tuning the linear layer on different dataset) in Sec.~\ref{ssec:results-comp-baselines}.
Our experiments utilized five different datasets, including CIFAR-10~\cite{cifar}, CIFAR-100~\cite{cifar}, STL-10~\cite{stl10}, TinyImageNet~\cite{tinyimagenet}, and ImageNet~\cite{ImageNet}.

\subsection{Implementation details}
For four datasets (CIFAR-10, CIFAR-100, STL-10, TinyImageNet), we used a modified ResNet-18~\cite{ResNet} as proposed by \citet{ReSSL}.
We basically followed the experimental settings from ReSSL~\cite{ReSSL}.
To compare with the baselines, we pretrained networks for 200 epochs and tuned only a linear layer on top of the frozen pretrained backbone model for 100 epochs (linear probing).
Keep in mind that our intention is to examine the three effects on SSL methods rather than to compare the performance of different SSL methods.
Therefore, we fixed the training hyper-parameters across different SSL methods.

For ImageNet, we used ResNet-50~\cite{ResNet} as the backbone model, following other SSL literature~\cite{MoCo,SimCLR,BYOL}.
We pretrained for 100 epochs and fine-tuned a linear layer for 90 epochs, consistent with previous works~\cite{MoCov3,SimCLR}.
We did not alter any hyperparameters of the baselines, except for newly introduced hyperparameters ($s$ and $c$).
Please refer to the appendix for more detailed experimental settings due to page limits.

\begin{figure*}[ht]
\centering
\includegraphics[width=0.95\linewidth]{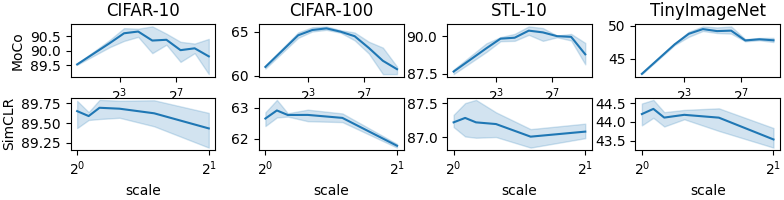}
\vspace{-0.2cm}
\caption{
Top 1 accuracy when emphasizing positive samples using $s$ (Eq.~\ref{eq:pos-emp}).
}
\vspace{-0.25cm}
\label{fig:pos-emp}
\end{figure*}

\begin{figure}[ht]
\centering
\includegraphics[width=0.95\linewidth]{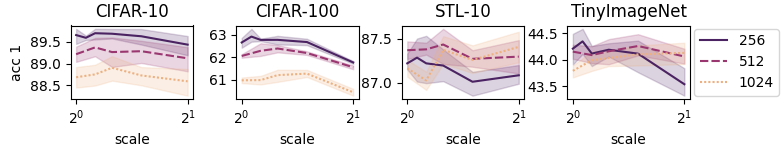}
\vspace{-0.5cm}
\caption{Relationship between the batch size and the positive sample gradient multiplier $s$ in SimCLR. The x axis is for $s$ and is in log scale.}
\label{fig:reb-simclr-bs}
\end{figure}

\subsection{Emphasizing positive samples}
\label{ssec:results-emphasizing-positive}
In this section, we quantitatively measure the effect of emphasizing positive samples.
To this end, we introduce a new hyper-parameter $s$ to scale positive sample gradients as follows:
\begin{equation}
    w_{ij} = (1-p_{ij}) + s \cdot p_{ij}
    \label{eq:pos-emp}
\end{equation}
\begin{equation}
    \delta_{ij}^{scale} = \delta_{ij}w_{ij} + sg(\delta_{ij}) \cdot (1-w_{ij}).
    \label{eq:pos-emp-overall}
\end{equation}
$w_{ij}$ denotes sample-wise weight, and $sg(\cdot)$ is the stop gradient operation.
We trained neural networks by replacing $\delta_{ij}$ with $\delta_{ij}^{scale}$ in Eq.~\ref{eq:generalized-eq-with-margin}.

We tested the effect of emphasizing positive samples on both MoCo and SimCLR.
BYOL was excluded because it only uses positive samples and thus emphasizing positive samples is the same as increasing the learning rate.
Fig.~\ref{fig:pos-emp} visualizes the results.

As the figure shows, scaling up positive sample gradients significantly improves the performance of MoCo.
It shows that lowering the relative weights of negative samples can improve the performance, but if they get too small, it will rather cause performance degradation.
Given that the gradients of MoCo approach those of BYOL as the scale $s$ approaches infinity (assuming the learning rate is adjusted accordingly), the presence of negative samples may be rather necessary for better SSL performance.
It is also worth noting that emphasizing positive samples yields higher accuracy than using margins in the case of MoCo (Tab.~\ref{tab:main_frozen}).
This implies that the performance improvement brought by emphasizing positive samples is offset by other factors, such as curvatures (Sec.~\ref{ssec:results-curvature}).

Unlike MoCo, SimCLR does not seem to improve by scaling up positive sample gradients.
However, Fig.~\ref{fig:reb-simclr-bs} shows that as the batch size increases, the optimal $s$ increases, as does the performance gap.
The larger the batch size, the more the performance curve becomes similar to that of MoCo.
We believe this is related to the issue that SimCLR requires a large batch size~\cite{SimCLR}.

In conclusion, while the peak and slopes are variable, it exhibits a consistent pattern across various methods and datasets.
Considering that many algorithms quickly converge in a few epochs, but there is no gain in much longer training epochs, we also include the experimental results of MoCo with the training epochs of 1,000 in the appendix.

\begin{figure*}[]
\centering
\includegraphics[width=0.95\linewidth]{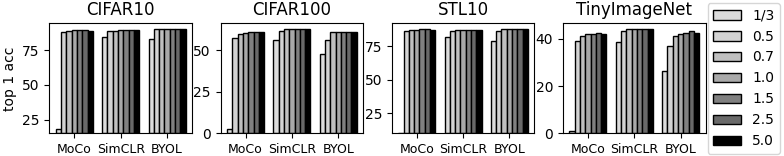}
\vspace{-0.3cm}
\caption{Controlling positive sample gradient scale curvatures using $c$ (Eq.~\ref{eq:curve-control}). $s$ was set to one.
}
\label{fig:curve-acc}
\end{figure*}

\begin{figure*}[]
    \centering
    \includegraphics[width=0.95\linewidth]{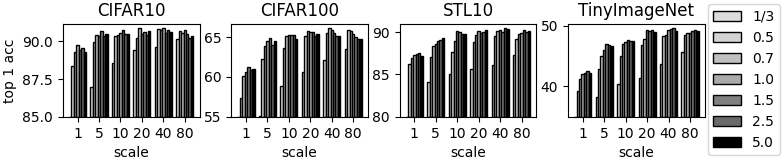}
    \vspace{-0.3cm}
    \caption{Controlling both the scale $s$ and curvature $c$ of MoCo (Eq.~\ref{eq:curve-control}).}
    \label{fig:sc-curve-acc}
    \vspace{-0.3cm}
\end{figure*}

\begin{figure}[]
\centering
\includegraphics[width=0.95\linewidth]{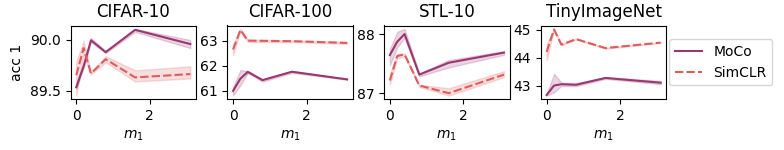}
\vspace{-0.5cm}
\caption{The 1 accuracy when scaling gradients by the ratio (Sec.~\ref{ssec:ratio}).}
\label{fig:third-ratio}
\vspace{-0.5cm}
\end{figure}

\subsection{Curvature of the positive sample gradient scale}
\label{ssec:results-curvature}
In this section, we analyze the effect of weighting positive samples differently based on their angles $p_{il}$.
We will refer to a weight curve, a function of the positive sample angle, as a curvature.
To experiment with convex, linear, and concave curvatures, we define a curve $\gamma(x, c)$ as follows:
\begin{align}
    \gamma(x, c) = |(1 - x^c)^{1/c}|,
    \label{eq:curves}
\end{align}
where $c$ is a parameter controlling the curvature, and $|\cdot|$ denotes the absolute operation.
When $c$ is infinity, it becomes exactly the same as not controlling the curvature at all.
To use Eq.~\ref{eq:curves} to control the diminishing rate of the positive samples gradient, we set the weights as follows;
\begin{align}
    w_{ij}^{dim} = (1-p_{ij}) + \gamma(\theta_{ij}/\pi, c) \cdot s \cdot p_{ij}.
    \label{eq:curve-control}
\end{align}
For experiments, we replaced $w_{ij}$ with $w_{ij}^{dim}$ in Eq.~\ref{eq:pos-emp-overall}.
We used three convex ($c$ is 1/3, 0.5, 0.7), three concave ($c$ is 1.5, 2.5, 5), and linear ($c$ is 1) curves.

Fig.~\ref{fig:curve-acc} shows how different $c$ affects various SSL methods and datasets, and Fig.~\ref{fig:sc-curve-acc} shows the performance of MoCo as both scale $s$ and $c$ change.
If $s$ is 1, the performance variation caused by $c$ is negligible, unless $c$ is extremely small.
However, the performance variation caused by $c$ becomes more pronounced as $s$ increases.
While it varies slightly depending on the dataset, in many cases, $c$ being close to or greater than 1 yields the optimal outcome.
Furthermore, considering that the curvature due to margins is highly convex, this explains partly why using margins might not fully exploit the advantage of weighting positive samples differently based on their angles.

\begin{table}[ht]
    \centering
    \caption{The effect of each gradient attenuation type and attenuation magnitude $\alpha$.}
    \begin{tabular}{llcccccccc}
    \toprule
    \multicolumn{2}{l}{} & \multicolumn{2}{c}{CIFAR10} & \multicolumn{2}{c}{CIFAR100} & \multicolumn{2}{c}{STL10} & \multicolumn{2}{c}{TinyImageNet} \\
    \cmidrule(r){3-4} \cmidrule(r){5-6} \cmidrule(r){7-8}\cmidrule(r){9-10}
    method & $\alpha$ & type I & type II & type I & type II & type I & type II & type I & type II\\
    \midrule
    MoCo & 0 & \multicolumn{2}{c}{89.413} & \multicolumn{2}{c}{60.956} & \multicolumn{2}{c}{87.629} & \multicolumn{2}{c}{42.133} \\
     & 0.25 & 89.376 & 89.713 & 60.857 & 61.217 & 87.292 & 87.455 & 42.420 & 42.460 \\
     & 1 & 89.743 & 89.633 & 60.997 & 61.263 & 87.329 & 87.442 & 42.797 & 42.510 \\
    \midrule
    SimCLR & 0 & \multicolumn{2}{c}{89.653} & \multicolumn{2}{c}{62.655} & \multicolumn{2}{c}{87.188} & \multicolumn{2}{c}{44.184}\\
     & 0.25 & 89.867 & 89.557 & 62.407 & 62.763 & 87.200 & 87.042 & 44.206 & 44.197\\
     & 1 & 89.735 & 89.767 & 62.840 & 62.800 & 87.300 & 87.317 & 44.260 & 44.260 \\
    \bottomrule
    \end{tabular}
    \label{tab:attenuation}
\end{table}

\begin{table}[ht]
\centering
\caption{Linear probing. \textit{pos.} denotes emphasizing positive samples and \textit{curv.} denotes controlling the curvature of positive gradient scales.}
\begin{tabular}{lcccc}
    \toprule
     & CIFAR10 & CIFAR100 & STL10 & TinyImageNet \\
     \midrule
     MoCo & 89.413 $\pm$ 0.109 & 60.956 $\pm$ 0.167 & 87.629 $\pm$ 0.175 & 42.133 $\pm$ 0.225 \\
     \textit{\ \ \ \ \ \ \ \ + margins} & 89.840 $\pm$ 0.105 & 61.796 $\pm$ 0.254 & 87.688 $\pm$ 0.275 & 42.860 $\pm$ 0.280 \\
     \textit{\ \ \ \ \ \ \ \ + pos.} & 90.663 $\pm$ 0.159 & 65.413 $\pm$ 0.205 & 90.357 $\pm$ 0.398 & 48.273 $\pm$ 0.405 \\
    \textit{\ \ \ \ \ \ \ \ + pos. \& curv.} & \textbf{90.865 $\pm$ 0.025} & \textbf{66.095 $\pm$ 0.295} & \textbf{90.361 $\pm$ 0.057} & \textbf{48.503 $\pm$ 0.270} \\
    \textit{\ \ \ \ \ \ \ \ + ratio.} & 90.100 $\pm$ 0.017 & 61.767 $\pm$ 0.040 & 88.000 $\pm$ 0.078 & 43.283 $\pm$ 0.031 \\ 
    \midrule
    SimCLR & 89.653 $\pm$ 0.207 & 62.655 $\pm$ 0.277 & 87.188 $\pm$ 0.145  & 44.184 $\pm$ 0.220 \\
     \textit{\ \ \ \ \ \ \ \ + margins} & \textbf{90.447 $\pm$ 0.191} & \textbf{63.507 $\pm$ 0.414} & 87.430 $\pm$ 0.220 & 44.593 $\pm$ 0.410 \\
     \textit{\ \ \ \ \ \ \ \ + pos.} & 89.695 $\pm$ 0.150 & 62.917 $\pm$ 0.302 & 87.284 $\pm$ 0.248 & 44.348 $\pm$ 0.295 \\
     \textit{\ \ \ \ \ \ \ \ + pos. \& curv.} & 90.190 $\pm$ 0.067 & 63.503 $\pm$ 0.279 & 87.346 $\pm$ 0.156 & 44.376 $\pm$ 0.201 \\
     \textit{\ \ \ \ \ \ \ \ + ratio.} & 89.920 $\pm$ 0.066 & 63.437 $\pm$ 0.045 & \textbf{87.658 $\pm$ 0.029} & \textbf{45.010 $\pm$ 0.040} \\ 

    \midrule 
    BYOL & 90.283 $\pm$ 0.109  & 61.006 $\pm$ 0.140 & 87.546 $\pm$ 0.668 & 41.846 $\pm$ 0.097 \\ 
     \textit{\ \ \ \ \ \ \ \ + curv.} & \textbf{90.485 $\pm$ 0.085} & \textbf{61.170 $\pm$ 0.340} & \textbf{88.051 $\pm$ 0.230} & \textbf{43.332 $\pm$ 0.564} \\ 
    \bottomrule
\end{tabular}
\label{tab:main_frozen}
\end{table}

\begin{table}[ht]
\centering
\caption{Transfer learning.
While freezing the pretrained backbone model, only the linear layer was tuned to the target dataset.
It is expressed in the form of ``pretrained dataset $\xrightarrow{}$ target dataset".}
\begin{tabular}{lcccc}
    \toprule
     & CIFAR100 & CIFAR10 & TinyImageNet & STL10 \\
     & $\xrightarrow{}$ CIFAR10 & $\xrightarrow{}$ CIFAR100 & $\xrightarrow{}$ STL10 & $\xrightarrow{}$ TinyImageNet \\
     \midrule
     MoCo & 75.417 $\pm$ 0.241 & 43.590 $\pm$ 0.225 & 72.013 $\pm$ 0.250 & 31.140 $\pm$ 0.130 \\ 
     \textit{\ \ \ \ \ \ \ \ + margins} & 75.380 $\pm$ 0.418 & 43.190 $\pm$ 0.255 & 72.154 $\pm$ 0.654 & 30.543 $\pm$ 0.160 \\ 
     \textit{\ \ \ \ \ \ \ \ + pos.} & \textbf{78.847 $\pm$ 0.244} & \textbf{55.113 $\pm$ 0.434} & 78.100 $\pm$ 0.185 & \textbf{41.825 $\pm$ 0.431} \\ 
     \textit{\ \ \ \ \ \ \ \ + pos. \& curv.} & 78.600 $\pm$ 0.042 & 52.940 $\pm$ 0.750 & \textbf{78.734 $\pm$ 0.040} & 41.777 $\pm$ 0.690 \\ 
     \textit{\ \ \ \ \ \ \ \ + ratio.} & 76.027 $\pm$ 0.074 & 44.113 $\pm$ 0.012 & 72.567 $\pm$ 0.040 & 32.070 $\pm$ 0.050 \\
    \midrule
    SimCLR & 77.300 $\pm$ 0.014 & 48.845 $\pm$ 0.615 & 75.533 $\pm$ 0.026 & 35.108 $\pm$ 0.453 \\ 
     \textit{\ \ \ \ \ \ \ \ + margins} & 76.377 $\pm$ 0.301 & 47.023 $\pm$ 0.219 & 75.617 $\pm$ 0.273 & 34.747 $\pm$ 0.430 \\
     \textit{\ \ \ \ \ \ \ \ + pos.} & 76.893 $\pm$ 0.220 & 49.090 $\pm$ 0.262 & 75.675 $\pm$ 0.194 & 35.067 $\pm$ 0.380 \\ 
     \textit{\ \ \ \ \ \ \ \ + pos. \& curv.} & 76.620 $\pm$ 0.190 & 48.683 $\pm$ 0.107 & 75.571 $\pm$ 0.220  & 34.280 $\pm$ 0.215 \\ 
    \textit{\ \ \ \ \ \ \ \ + ratio.} & \textbf{77.440 $\pm$ 0.346} & \textbf{49.903 $\pm$ 0.025} & \textbf{75.983 $\pm$ 0.029} & \textbf{35.237 $\pm$ 0.032} \\ 

    \midrule
    BYOL & \textbf{75.086 $\pm$ 0.156} & \textbf{41.560 $\pm$ 0.321} & 69.433 $\pm$ 0.272 & \textbf{30.306 $\pm$ 0.309} \\ 
     \textit{\ \ \ \ \ \ \ \ + curv.} & 74.820 $\pm$ 0.552 & 39.445 $\pm$ 0.078 & \textbf{72.442 $\pm$ 0.366} & 29.260 $\pm$ 0.028 \\ 
    \bottomrule
\end{tabular}
\label{tab:main-transfer}
\end{table}

\begin{table}[h] 
    \begin{center}
    \captionof{table}{Top 1 accuracy of ResNet-50 when linear probing on ImageNet.}
    \begin{tabular}{lcc}
    \toprule
    & Epoch & ImageNet \\
    \midrule
    MoCov3 & 100 & 68.9 \\
    \textit{\quad + pos. \& curv.} & 100 & \textbf{70.9} \\
    \midrule
    SimCLR & 100 & 64.7 \\
    \textit{\quad + pos. \& curv.} & 100 & \textbf{65.7} \\ 
    \bottomrule
    \end{tabular}
    \label{tab:imagenet}
    \end{center}
\end{table}

\subsection{The ratio of sums of exponentiated logits}
\label{ssec:ratio}
We also conducted experiments on the effect of the ratio $(p_{ij} - \beta q_{ij})/ (p_{ij} - \beta \tilde{q}_{ij})$.
To this end, we only multiply the gradients by the ratio without modifying the objective function.
That is, we used the loss without margins (Eq.~\ref{eq:generalized-eq}), while scaling gradients by the ratio.
For experiments, we calculated the ratios using $m_1$ of 0.2, 0.4, 0.8, 1.6, and 3.1.
Fig.~\ref{fig:third-ratio} shows that scaling the gradients by the ratio can improve performance.
Usually, the optimal margin $m_1$ lies somewhere larger than zero.
But the performance curves show less congruent patterns.

\subsection{Attenuating the diminishing gradients}
\label{ssec:results-attenuation}
As explained in Secs.~\ref{sssec:ga-angular-margin} and \ref{ssec:ga-subtractive-margin}, margins can attenuate the diminishing gradients as $q_{ij}$ approaches $p_{ij}$.
Since the scale depends only on the positive sample, we ran experiments on two types of attenuation scales; type I and II.
They are defined as follows:
\begin{align}
    w_{ij}^{\text{I}} = \sum_{k \in \mathcal{X}} \frac{p_{ik}}{1 - \alpha \tilde{q}_{ik}},
    \quad 
    w_{ij}^{\text{II}} = (1 - p_{ij}) + \frac{p_{ij}}{p_{ij} - \alpha \tilde{q}_{ij}}.
    \label{eq:type1}
\end{align}
For type I, we multiply $1 / (1 - \alpha \tilde{q}_{il})$ to gradients to both positive and negative samples, as $m_2$ does (Eq.~\ref{eq:gradient-analysis-m2}).
If $\alpha$ equals $1-\exp((\cos(\theta_{il} + m_1) - \cos(\theta_{il})-m_2)/\tau)$, type I equals using subtractive margin $m_2$ (Eq.~\ref{eq:gradient-analysis-m2}).
For type II, we only scale gradients of positive samples only.

Tab.~\ref{tab:attenuation} shows the performance as $\alpha$ changes.
We exempt BYOL because $q_{ij}$ cannot exist without negative samples.
As the table shows, attenuating positive gradients as $q_{il}$ approaches $p_{il}$ does not significantly improve performance in both cases (types I and II). 
This might be related to the fact that many face recognition methods use only the angular margin $m_1$.

\subsection{Comparison with baselines}
\label{ssec:results-comp-baselines}
In this section, we compare the baselines with and without margins as well as three other components; emphasizing positive samples (in short \textit{pos.}), weighting positive samples differently (abbreviated as \textit{curv.}), and scaling by ratios (in short \textit{ratio.}).
We tested each model in two different settings; linear probing (Tab.~\ref{tab:main_frozen}) and transfer learning (Tab.~\ref{tab:main-transfer}).
Experimental details, including exact values of $s$ and $c$, are addressed in the appendix.

Tabs.~\ref{tab:main_frozen} and \ref{tab:main-transfer} present the linear probing and transfer learning performance on four datasets.
In addition, Tab.~\ref{tab:imagenet} shows the linear probing performance on ImageNet.
As indicated in Tab.~\ref{tab:main_frozen}, the most significant improvement of MoCo occurs when positive samples are emphasized.
The transfer learning performance (Tab.~\ref{tab:main-transfer}) demonstrates that this performance improvement is not due to overfitting but rather reflects enhanced representations capable of generalizing to unseen datasets.
While $curv.$ coupled with $pos.$ can improve performance in seen datasets, it does not consistently enhance performance in unseen datasets.
As mentioned in Sec.~\ref{ssec:results-emphasizing-positive}, emphasizing positive samples does not significantly improve SimCLR on four datasets.
On ImageNet (Tab.~\ref{tab:imagenet}), however, SimCLR shows improvement, and we believe this pertains to the increased batch size. 

Scaling gradients by the ratio consistently improves contrastive learning across various datasets, not only in seen datasets but also in unseen datasets.
The performance improvement is more apparent for SimCLR.
BYOL is structurally less affected by margins or their associated effects.
Consequently, performance improvements are limited.
In conclusion, optimizing margins and related effects can contribute to performance enhancements on the target dataset, and emphasizing positive samples and scaling by the ratio appear to be important for achieving improved representations that generalize to unseen datasets.

\section{Limitations and Discussion}
\label{sec:limit}
Our work is based on several assumptions; cosine similarity and one-hot target probability $p_{ij}$.
Since not all methods follow these assumptions, our work requires further validation to determine whether the observations made in this study can be transferred to other contrastive learning methods that violate these assumptions~\cite{debiased, ReSSL}.
Moreover, we could not delve into the ratio $(p_{ij} - \beta q_{ij})/ (p_{ij} - \beta \tilde{q}_{ij})$ due to the fact that it is not a separable function.
Considering the improvement in SimCLR, delving into it can be an interesting direction.

\section{Conclusion}
\label{sec:conclusion}
We proposed a novel view on understanding the role of margins using gradient analysis and not relying on decision-boundary-based explanations.
By analyzing gradients, we discovered that margins have a mixture of four different effects: emphasizing positive samples, weighting positive samples differently based on their angles, scaling gradients by the ratio of sums of exponentiated logits, and alleviating the diminishing gradient effect as the estimated probability approaches the target probability.
We separated each effect and experimentally demonstrated the significance and limits of each.
We hope our analysis of how margins affect the gradients of representation learning will help improve contrastive learning and possibly margins themselves.

\bibliography{iclr2024_conference}

\begin{thebibliography}{33}
\providecommand{\natexlab}[1]{#1}
\providecommand{\url}[1]{\texttt{#1}}
\expandafter\ifx\csname urlstyle\endcsname\relax
  \providecommand{\doi}[1]{doi: #1}\else
  \providecommand{\doi}{doi: \begingroup \urlstyle{rm}\Url}\fi

\bibitem[Boutros et~al.(2022)Boutros, Damer, Kirchbuchner, and
  Kuijper]{ElasticFace}
Fadi Boutros, Naser Damer, Florian Kirchbuchner, and Arjan Kuijper.
\newblock Elasticface: Elastic margin loss for deep face recognition.
\newblock In \emph{Proceedings of the IEEE/CVF Conference on Computer Vision
  and Pattern Recognition (CVPR) Workshops}, pp.\  1578--1587, 6 2022.

\bibitem[Caron et~al.(2020)Caron, Misra, Mairal, Goyal, Bojanowski, and
  Joulin]{SwAV}
Mathilde Caron, Ishan Misra, Julien Mairal, Priya Goyal, Piotr Bojanowski, and
  Armand Joulin.
\newblock Unsupervised learning of visual features by contrasting cluster
  assignments.
\newblock In H~Larochelle, M~Ranzato, R~Hadsell, M~F Balcan, and H~Lin (eds.),
  \emph{Advances in Neural Information Processing Systems}, volume~33, pp.\
  9912--9924. Curran Associates, Inc., 2020.
\newblock SwAV.

\bibitem[Chen et~al.(2020{\natexlab{a}})Chen, Kornblith, Norouzi, and
  Hinton]{SimCLR}
Ting Chen, Simon Kornblith, Mohammad Norouzi, and Geoffrey Hinton.
\newblock A simple framework for contrastive learning of visual
  representations.
\newblock In Hal~Daumé III and Aarti Singh (eds.), \emph{Proceedings of the
  37th International Conference on Machine Learning}, volume 119, pp.\
  1597--1607. PMLR, 9 2020{\natexlab{a}}.
\newblock SimCLR.

\bibitem[Chen et~al.(2020{\natexlab{b}})Chen, Kornblith, Swersky, Norouzi, and
  Hinton]{SimCLRv2}
Ting Chen, Simon Kornblith, Kevin Swersky, Mohammad Norouzi, and Geoffrey~E
  Hinton.
\newblock Big self-supervised models are strong semi-supervised learners.
\newblock In H~Larochelle, M~Ranzato, R~Hadsell, M~F Balcan, and H~Lin (eds.),
  \emph{Advances in Neural Information Processing Systems}, volume~33, pp.\
  22243--22255. Curran Associates, Inc., 2020{\natexlab{b}}.
\newblock SimCLRv2.

\bibitem[Chen \& He(2021)Chen and He]{SimSIAM}
Xinlei Chen and Kaiming He.
\newblock Exploring simple siamese representation learning.
\newblock In \emph{Proceedings of the IEEE/CVF Conference on Computer Vision
  and Pattern Recognition (CVPR)}, pp.\  15750--15758, 6 2021.
\newblock SimSIAM.

\bibitem[Chen et~al.(2020{\natexlab{c}})Chen, Fan, Girshick, and He]{MoCov2}
Xinlei Chen, Haoqi Fan, Ross Girshick, and Kaiming He.
\newblock Improved baselines with momentum contrastive learning.
\newblock \emph{arXiv preprint arXiv:2003.04297}, 2020{\natexlab{c}}.

\bibitem[Chen et~al.(2021)Chen, Xie, and He]{MoCov3}
Xinlei Chen, Saining Xie, and Kaiming He.
\newblock An empirical study of training self-supervised vision transformers.
\newblock In \emph{Proceedings of the IEEE/CVF International Conference on
  Computer Vision (ICCV)}, pp.\  9640--9649, 10 2021.

\bibitem[Chuang et~al.(2020)Chuang, Robinson, Lin, Torralba, and
  Jegelka]{debiased}
Ching-Yao Chuang, Joshua Robinson, Yen-Chen Lin, Antonio Torralba, and Stefanie
  Jegelka.
\newblock Debiased contrastive learning.
\newblock In H.~Larochelle, M.~Ranzato, R.~Hadsell, M.F. Balcan, and H.~Lin
  (eds.), \emph{Advances in Neural Information Processing Systems}, volume~33,
  pp.\  8765--8775. Curran Associates, Inc., 2020.

\bibitem[Coates et~al.(2011)Coates, Ng, and Lee]{stl10}
Adam Coates, Andrew Ng, and Honglak Lee.
\newblock An analysis of single-layer networks in unsupervised feature
  learning.
\newblock In \emph{Proceedings of the fourteenth international conference on
  artificial intelligence and statistics}, pp.\  215--223. JMLR Workshop and
  Conference Proceedings, 2011.

\bibitem[Deng et~al.(2019)Deng, Guo, Xue, and Zafeiriou]{ArcFace}
Jiankang Deng, Jia Guo, Niannan Xue, and Stefanos Zafeiriou.
\newblock Arcface: Additive angular margin loss for deep face recognition.
\newblock In \emph{Proceedings of the IEEE/CVF conference on computer vision
  and pattern recognition}, pp.\  4690--4699, 2019.

\bibitem[Desplanques et~al.(2020)Desplanques, Thienpondt, and
  Demuynck]{ECAPA-TDNN}
Brecht Desplanques, Jenthe Thienpondt, and Kris Demuynck.
\newblock {ECAPA-TDNN: Emphasized Channel Attention, Propagation and
  Aggregation in TDNN Based Speaker Verification}.
\newblock In \emph{Proc. Interspeech 2020}, pp.\  3830--3834, 2020.
\newblock \doi{10.21437/Interspeech.2020-2650}.

\bibitem[Devlin et~al.(2019)Devlin, Chang, Lee, and Toutanova]{devlin2018bert}
Jacob Devlin, Ming{-}Wei Chang, Kenton Lee, and Kristina Toutanova.
\newblock {BERT:} pre-training of deep bidirectional transformers for language
  understanding.
\newblock In \emph{Proceedings of the 2019 Conference of the North American
  Chapter of the Association for Computational Linguistics: Human Language
  Technologies, {NAACL-HLT} 2019, Minneapolis, MN, USA, June 2-7, 2019, Volume
  1 (Long and Short Papers)}, pp.\  4171--4186. Association for Computational
  Linguistics, 2019.

\bibitem[Dosovitskiy et~al.(2021)Dosovitskiy, Beyer, Kolesnikov, Weissenborn,
  Zhai, Unterthiner, Dehghani, Minderer, Heigold, Gelly, Uszkoreit, and
  Houlsby]{dosovitskiy2020image}
Alexey Dosovitskiy, Lucas Beyer, Alexander Kolesnikov, Dirk Weissenborn,
  Xiaohua Zhai, Thomas Unterthiner, Mostafa Dehghani, Matthias Minderer, Georg
  Heigold, Sylvain Gelly, Jakob Uszkoreit, and Neil Houlsby.
\newblock An image is worth 16x16 words: Transformers for image recognition at
  scale.
\newblock In \emph{International Conference on Learning Representations}, 2021.

\bibitem[Grill et~al.(2020)Grill, Strub, Altché, Tallec, Richemond,
  Buchatskaya, Doersch, Pires, Guo, Azar, Piot, koray kavukcuoglu, Munos, and
  Valko]{BYOL}
Jean-Bastien Grill, Florian Strub, Florent Altché, Corentin Tallec, Pierre
  Richemond, Elena Buchatskaya, Carl Doersch, Bernardo~Avila Pires, Zhaohan
  Guo, Mohammad~Gheshlaghi Azar, Bilal Piot, koray kavukcuoglu, Remi Munos, and
  Michal Valko.
\newblock Bootstrap your own latent - a new approach to self-supervised
  learning.
\newblock In H~Larochelle, M~Ranzato, R~Hadsell, M~F Balcan, and H~Lin (eds.),
  \emph{Advances in Neural Information Processing Systems}, volume~33, pp.\
  21271--21284. Curran Associates, Inc., 2020.
\newblock BYOL.

\bibitem[HaoChen et~al.(2021)HaoChen, Wei, Gaidon, and
  Ma]{NEURIPS2021_27debb43}
Jeff~Z. HaoChen, Colin Wei, Adrien Gaidon, and Tengyu Ma.
\newblock Provable guarantees for self-supervised deep learning with spectral
  contrastive loss.
\newblock In M.~Ranzato, A.~Beygelzimer, Y.~Dauphin, P.S. Liang, and J.~Wortman
  Vaughan (eds.), \emph{Advances in Neural Information Processing Systems},
  volume~34, pp.\  5000--5011. Curran Associates, Inc., 2021.
\newblock URL
  \url{https://proceedings.neurips.cc/paper_files/paper/2021/file/27debb435021eb68b3965290b5e24c49-Paper.pdf}.

\bibitem[He et~al.(2016)He, Zhang, Ren, and Sun]{ResNet}
Kaiming He, Xiangyu Zhang, Shaoqing Ren, and Jian Sun.
\newblock Deep residual learning for image recognition.
\newblock In \emph{Proceedings of the IEEE Conference on Computer Vision and
  Pattern Recognition (CVPR)}, June 2016.

\bibitem[He et~al.(2020)He, Fan, Wu, Xie, and Girshick]{MoCo}
Kaiming He, Haoqi Fan, Yuxin Wu, Saining Xie, and Ross Girshick.
\newblock Momentum contrast for unsupervised visual representation learning.
\newblock In \emph{Proceedings of the IEEE/CVF Conference on Computer Vision
  and Pattern Recognition (CVPR)}, 6 2020.
\newblock MoCo.

\bibitem[Hsu et~al.(2021)Hsu, Bolte, Tsai, Lakhotia, Salakhutdinov, and
  Mohamed]{hsu2021hubert}
Wei-Ning Hsu, Benjamin Bolte, Yao-Hung~Hubert Tsai, Kushal Lakhotia, Ruslan
  Salakhutdinov, and Abdelrahman Mohamed.
\newblock Hubert: Self-supervised speech representation learning by masked
  prediction of hidden units.
\newblock \emph{IEEE/ACM Transactions on Audio, Speech, and Language
  Processing}, 29:\penalty0 3451--3460, 2021.

\bibitem[Huang et~al.(2020)Huang, Wang, Tai, Liu, Shen, Li, Li, and
  Huang]{CurricularFace}
Yuge Huang, Yuhan Wang, Ying Tai, Xiaoming Liu, Pengcheng Shen, Shaoxin Li,
  Jilin Li, and Feiyue Huang.
\newblock Curricularface: Adaptive curriculum learning loss for deep face
  recognition.
\newblock In \emph{Proceedings of the IEEE/CVF Conference on Computer Vision
  and Pattern Recognition (CVPR)}, 6 2020.

\bibitem[Kim et~al.(2022)Kim, Jain, and Liu]{AdaFace}
Minchul Kim, Anil~K Jain, and Xiaoming Liu.
\newblock Adaface: Quality adaptive margin for face recognition.
\newblock In \emph{Proceedings of the IEEE/CVF Conference on Computer Vision
  and Pattern Recognition (CVPR)}, pp.\  18750--18759, 6 2022.

\bibitem[Krizhevsky et~al.(2009)Krizhevsky, Hinton, et~al.]{cifar}
Alex Krizhevsky, Geoffrey Hinton, et~al.
\newblock Learning multiple layers of features from tiny images, 2009.

\bibitem[Le \& Yang(2015)Le and Yang]{tinyimagenet}
Ya~Le and Xuan Yang.
\newblock Tiny imagenet visual recognition challenge.
\newblock \emph{CS 231N}, 7\penalty0 (7):\penalty0 3, 2015.

\bibitem[Meng et~al.(2021)Meng, Zhao, Huang, and Zhou]{MagFace}
Qiang Meng, Shichao Zhao, Zhida Huang, and Feng Zhou.
\newblock Magface: A universal representation for face recognition and quality
  assessment.
\newblock In \emph{Proceedings of the IEEE/CVF Conference on Computer Vision
  and Pattern Recognition (CVPR)}, pp.\  14225--14234, 6 2021.

\bibitem[Oord et~al.(2018)Oord, Li, and Vinyals]{infoNCE(CPC)}
Aaron van~den Oord, Yazhe Li, and Oriol Vinyals.
\newblock Representation learning with contrastive predictive coding.
\newblock \emph{arXiv preprint arXiv:1807.03748}, 2018.

\bibitem[Russakovsky et~al.(2015)Russakovsky, Deng, Su, Krause, Satheesh, Ma,
  Huang, Karpathy, Khosla, Bernstein, Berg, and Fei-Fei]{ImageNet}
Olga Russakovsky, Jia Deng, Hao Su, Jonathan Krause, Sanjeev Satheesh, Sean Ma,
  Zhiheng Huang, Andrej Karpathy, Aditya Khosla, Michael Bernstein,
  Alexander~C. Berg, and Li~Fei-Fei.
\newblock Imagenet large scale visual recognition challenge.
\newblock \emph{Int. J. Comput. Vision}, 115\penalty0 (3):\penalty0 211–252,
  dec 2015.
\newblock ISSN 0920-5691.
\newblock \doi{10.1007/s11263-015-0816-y}.
\newblock URL \url{https://doi.org/10.1007/s11263-015-0816-y}.

\bibitem[Wang \& Liu(2021)Wang and Liu]{Wang_2021_CVPR}
Feng Wang and Huaping Liu.
\newblock Understanding the behaviour of contrastive loss.
\newblock In \emph{Proceedings of the IEEE/CVF Conference on Computer Vision
  and Pattern Recognition (CVPR)}, pp.\  2495--2504, June 2021.

\bibitem[Wang et~al.(2018)Wang, Wang, Zhou, Ji, Gong, Zhou, Li, and
  Liu]{CosFace}
Hao Wang, Yitong Wang, Zheng Zhou, Xing Ji, Dihong Gong, Jingchao Zhou, Zhifeng
  Li, and Wei Liu.
\newblock Cosface: Large margin cosine loss for deep face recognition.
\newblock In \emph{Proceedings of the IEEE Conference on Computer Vision and
  Pattern Recognition (CVPR)}, 6 2018.

\bibitem[Wang et~al.(2022)Wang, Guo, Deng, and Lu]{Wang2022}
Haoqing Wang, Xun Guo, Zhi-Hong Deng, and Yan Lu.
\newblock Rethinking minimal sufficient representation in contrastive learning.
\newblock In \emph{Proceedings of the IEEE/CVF Conference on Computer Vision
  and Pattern Recognition (CVPR)}, pp.\  16041--16050, 6 2022.

\bibitem[Wu et~al.(2018)Wu, Xiong, Yu, and Lin]{wu2018unsupervised}
Zhirong Wu, Yuanjun Xiong, Stella~X Yu, and Dahua Lin.
\newblock Unsupervised feature learning via non-parametric instance
  discrimination.
\newblock In \emph{Proceedings of the IEEE conference on computer vision and
  pattern recognition}, pp.\  3733--3742, 2018.

\bibitem[Zhan et~al.(2022)Zhan, Yu, Wu, Zhang, Lu, and Zhang]{Zhan2022}
Fangneng Zhan, Yingchen Yu, Rongliang Wu, Jiahui Zhang, Shijian Lu, and
  Changgong Zhang.
\newblock Marginal contrastive correspondence for guided image generation.
\newblock In \emph{Proceedings of the IEEE/CVF Conference on Computer Vision
  and Pattern Recognition (CVPR)}, pp.\  10663--10672, 6 2022.

\bibitem[Zhang et~al.(2022)Zhang, Ma, Wang, and Chua]{zhang2022incorporating}
An~Zhang, Wenchang Ma, Xiang Wang, and Tat-Seng Chua.
\newblock Incorporating bias-aware margins into contrastive loss for
  collaborative filtering.
\newblock In Alice~H. Oh, Alekh Agarwal, Danielle Belgrave, and Kyunghyun Cho
  (eds.), \emph{Advances in Neural Information Processing Systems}, 2022.
\newblock URL \url{https://openreview.net/forum?id=apC354ZsGwK}.

\bibitem[Zhang et~al.(2019)Zhang, Zhao, Qiao, Wang, and Li]{AdaCos}
Xiao Zhang, Rui Zhao, Yu~Qiao, Xiaogang Wang, and Hongsheng Li.
\newblock Adacos: Adaptively scaling cosine logits for effectively learning
  deep face representations.
\newblock In \emph{Proceedings of the IEEE/CVF Conference on Computer Vision
  and Pattern Recognition (CVPR)}, 6 2019.

\bibitem[Zheng et~al.(2021)Zheng, You, Wang, Qian, Zhang, Wang, and Xu]{ReSSL}
Mingkai Zheng, Shan You, Fei Wang, Chen Qian, Changshui Zhang, Xiaogang Wang,
  and Chang Xu.
\newblock Ressl: Relational self-supervised learning with weak augmentation.
\newblock In M~Ranzato, A~Beygelzimer, Y~Dauphin, P~S Liang, and J~Wortman
  Vaughan (eds.), \emph{Advances in Neural Information Processing Systems},
  volume~34, pp.\  2543--2555. Curran Associates, Inc., 2021.

\end{thebibliography}
\bibliographystyle{iclr2024_conference}

\clearpage

\appendix

\renewcommand\thesection{\Alph{section}}
\setcounter{section}{0}

\section{Proofs}

\subsection{Theorem \ref{theorem:main}}

Recalling that $\tilde{\mathcal{L}}_i = -\sum_{j \in \mathcal{X}}{p_{ij}\tilde{\delta}_{ij}} + \beta \sum_{j \in \mathcal{X}}{p_{ij}\log \sum_{k \in \mathcal{X}}{\exp(\tilde{\delta}_{ik})}}$, the derivative with respect to $\tilde{\delta}_{ij}$ can be expressed as follows:
\begin{equation}
    \frac{\partial \tilde{\mathcal{L}}_i}{\partial \tilde{\delta}_{ij}} = -p_{ij} + \beta \frac{\exp(\tilde{\delta}_{ij})}{\sum_{k \in \mathcal{X}}{\exp(\tilde{\delta}_{ik})}} = - p_{ij} + \beta \tilde{q}_{ij}.
\end{equation}
Similarly, $\partial \tilde{\mathcal{L}}_i / \partial \delta_{ij}$ equals $-p_{ij} + \beta q_{ij}$.
The derivative of $\tilde{\mathcal{L}}_i$ (Eq.~\ref{eq:generalized-eq}) and $\mathcal{L}_i$ (Eq.~\ref{eq:generalized-eq-with-margin}) with respect to $\theta_{ij}$ can be represented as follows:
\begin{equation}
    \frac{\partial \tilde{\mathcal{L}}_i}{\partial \theta_{ij}} = \frac{\partial \tilde{\mathcal{L}}_i}{\partial \tilde{\delta}_{ij}} \frac{\partial \tilde{\delta}_{ij}}{\partial \theta_{ij}} = (p_{ij} - \beta \tilde{q}_{ij})\frac{\sin(\theta_{ij})}{\tau}
    \label{eq:tilde-Li/theta-ij}
\end{equation}
\begin{equation}
    \frac{\partial \mathcal{L}_i}{\partial \theta_{ij}} = \frac{\partial \mathcal{L}_i}{\partial \delta_{ij}} \frac{\partial \delta_{ij}}{\partial \theta_{ij}} = (p_{ij} - \beta q_{ij})\frac{\sin(\theta_{ij} + m_1p_{ij})}{\tau}
    \label{eq:Li/theta-ij}
\end{equation}

Using Eqs.~\ref{eq:tilde-Li/theta-ij} and \ref{eq:Li/theta-ij}, we can draw the relationships between two derivatives (Theorem~\ref{theorem:main}).
\begin{equation}
    \frac{\partial \mathcal{L}_i}{\partial \theta_{ij}} = (p_{ij} - \beta q_{ij})\frac{\sin(\theta_{ij} + m_1p_{ij})}{\tau}\frac{p_{ij} - \beta \tilde{q}_{ij}}{p_{ij} - \beta \tilde{q}_{ij}}\frac{\sin(\theta_{ij})}{\sin(\theta_{ij})}
\end{equation}
\begin{equation}
    = (p_{ij} - \beta \tilde{q}_{ij})\frac{\sin(\theta_{ij})}{\tau}\frac{p_{ij} - \beta q_{ij}}{p_{ij} - \beta \tilde{q}_{ij}}\frac{\sin(\theta_{ij} + m_1p_{ij})}{\sin(\theta_{ij})} = \frac{\partial \tilde{\mathcal{L}}_i}{\partial \tilde{\delta}_{ij}}\frac{p_{ij} - \beta q_{ij}}{p_{ij} - \beta \tilde{q}_{ij}}\frac{\sin(\theta_{ij} + m_1p_{ij})}{\sin(\theta_{ij})}
\end{equation}

\subsection{Lemma \ref{lemma:ratios}}

We first assume that $\beta$ is one and $p_{ij}$ is either zero or one.
Before proving the lemman, we first reformulate $\tilde{q}_{ij}$ and $q_{ij}$ as follows:
\begin{align}
    \tilde{q}_{ij} = \frac{\exp(\tilde{\delta}_{ij})}{ \sum_{k \in \mathcal{X}}{\exp(\tilde{\delta}_{ik})}}
    = \frac{\exp(\tilde{\delta}_{ij})}{\exp(\tilde{\delta}_{il}) / \tilde{q}_{il}}
\end{align}
\begin{align}
    q_{ij} = \frac{\exp(\delta_{ij})}{ \sum_{k \in \mathcal{X}}{\exp(\delta_{ik})}} = \frac{\exp(\delta_{ij})}{\exp(\delta_{il}) - \exp(\tilde{\delta}_{il}) + \sum_{k \in \mathcal{X}}{\exp(\tilde{\delta}_{ik})}} \\
    = \frac{\exp(\delta_{ij})}{\exp(\delta_{il}) - \exp(\tilde{\delta}_{il}) + \exp(\tilde{\delta}_{il}) / \tilde{q}_{il}} \\
    = \frac{\exp(\delta_{ij})}{\exp(\delta_{il})+ \exp(\tilde{\delta}_{il}) (1/\tilde{q}_{il} - 1)}.
\end{align}
Using these reformulations, we can express $(p_{ij} - \beta q_{ij}) / (p_{ij} - \beta \tilde{q}_{ij})$ as follows:
\begin{align}
    \frac{p_{ij} - \beta q_{ij}}{p_{ij} - \beta \tilde{q}_{ij}} = \frac{(p_{ij}(\exp(\delta_{il}) + \exp(\tilde{\delta}_{il})(1/\tilde{q}_{il} - 1)) - \beta \exp(\delta_{ij})) / (\exp(\delta_{il}) + \exp(\tilde{\delta}_{il})(1/\tilde{q}_{il} - 1))}{(p_{ij}\exp(\tilde{\delta}_{il})/\tilde{q}_{il} - \beta \exp(\tilde{\delta}_{ij})) / (\exp(\tilde{\delta}_{il})/\tilde{q}_{il})} \\
    = \frac{p_{ij}(\exp(\delta_{il}) + \exp(\tilde{\delta}_{il})(1/\tilde{q}_{il} - 1)) - \beta \exp(\delta_{ij})}{p_{ij}\exp(\tilde{\delta}_{il})/\tilde{q}_{il} - \beta \exp(\tilde{\delta}_{ij})}
    \frac{\exp(\tilde{\delta}_{il})/\tilde{q}_{il}}{\exp(\delta_{il}) + \exp(\tilde{\delta}_{il})(1/\tilde{q}_{il} - 1)}
\end{align}

For a positive sample $l$,
\begin{align}
    \frac{p_{il} - \beta q_{il}}{p_{il} - \beta \tilde{q}_{il}} = \frac{\exp(\delta_{il}) + \exp(\tilde{\delta}_{il})(1/\tilde{q}_{il} - 1) - \exp(\delta_{ij})}{\exp(\tilde{\delta}_{il})/\tilde{q}_{il} - \exp(\tilde{\delta}_{ij})}
    \frac{\exp(\tilde{\delta}_{il})/\tilde{q}_{il}}{\exp(\delta_{il}) + \exp(\tilde{\delta}_{il})(1/\tilde{q}_{il} - 1)} \\
    = \frac{\exp(\tilde{\delta}_{il})/\tilde{q}_{il} - \exp(\tilde{\delta}_{ij})}{\exp(\tilde{\delta}_{il})/\tilde{q}_{il} - \exp(\tilde{\delta}_{ij})}
    \frac{\exp(\tilde{\delta}_{il})/\tilde{q}_{il}}{\exp(\delta_{il}) + \exp(\tilde{\delta}_{il})(1/\tilde{q}_{il} - 1)} \\
    = \frac{\exp(\tilde{\delta}_{il})/\tilde{q}_{il}}{\exp(\delta_{il}) + \exp(\tilde{\delta}_{il})(1/\tilde{q}_{il} - 1)}.
\end{align}

Using the fact that $\delta_{ih}$ equals $\tilde{\delta}_{ih}$ for a negative sample $h$, we can get the following equation:
\begin{align}
    \frac{p_{ih} - \beta q_{ih}}{p_{ih} - \beta \tilde{q}_{ih}} = \frac{- \exp(\delta_{ih})}{- \exp(\tilde{\delta}_{ih})}
    \frac{\exp(\tilde{\delta}_{il})/\tilde{q}_{il}}{\exp(\delta_{il}) + \exp(\tilde{\delta}_{il})(1/\tilde{q}_{il} - 1)} = \frac{\exp(\tilde{\delta}_{il})/\tilde{q}_{il}}{\exp(\delta_{il}) + \exp(\tilde{\delta}_{il})(1/\tilde{q}_{il} - 1)}
\end{align}

That is, $(p_{ij} - \beta q_{ij}) / (p_{ij} - \beta \tilde{q}_{ij})$ equals $(\exp(\tilde{\delta}_{il})/q_{il}) \ /\ (\exp(\delta_{il}) + \exp(\tilde{\delta}_{il})(1/q_{il}-1))$, and it can also be expressed as 
$\sum \exp(\tilde{\delta}_{ij}) / \sum \exp(\delta_{ij})$.

\subsection{Lemma \ref{lemma:subtractive}}
\label{ssec:sub-margin-proof}
We first assume that there are only one positive sample and other samples are negative.
This assumption holds true for SSL methods we are analyzing.
In addition, we also assume that $\beta$ to one (which holds true for MoCo and SimCLR).
If $\beta$ is zero, then subtractive margin $m_2$ cannot work.
Eq.~\ref{eq:Li/theta-ij} can be rewritten as follows:
\begin{align}
    \frac{\partial \mathcal{L}_i}{\partial \theta_{ij}} = (p_{ij} - \beta q_{ij})\frac{\sin(\theta_{ij} + m_1p_{ij})}{\tau} = \frac{-\beta\exp(\delta_{ij}) + p_{ij}\sum_{k \in \mathcal{X}} \exp(\delta_{ik})}{\sum_{k \in \mathcal{X}} \exp(\delta_{ik})}\frac{\sin(\theta_{ij} + m_1p_{ij})}{\tau} \\
    = \frac{-\beta\exp(\delta_{ij}) + p_{ij}\sum_{k \in \mathcal{X}} \exp(\delta_{ik})}{\sum_{k \in \mathcal{X}} \exp(\tilde{\delta}_{ik})}\frac{\sum_{k \in \mathcal{X}} \exp(\tilde{\delta}_{ik})}{\sum_{k \in \mathcal{X}} \exp(\delta_{ik})}\frac{\sin(\theta_{ij} + m_1p_{ij})}{\tau}.
\end{align}

In addition, we reformualte $\exp{\delta_{ij}}$ as follows,
\begin{align}
    exp(\delta_{ij}) = \exp(\frac{\cos(\theta_{ij})}{\tau})\exp(\frac{\cos(\theta_{ij} + m_1 p_{ij}) - \cos(\theta_{ij}) - m_2p_{ij}}{\tau}).
\end{align}
\begin{align}
    = exp(\tilde{\delta}_{ij})\exp(\frac{\cos(\theta_{ij} + m_1 p_{ij}) - \cos(\theta_{ij}) - m_2p_{ij}}{\tau})
\end{align}
For brevity, we will use the notation $\eta_{ij}$ for $\exp((\cos(\theta_{ij} + m_1 p_{ij}) - \cos(\theta_{ij})-m_2p_{ij})/\tau)$.
We can rewrite the derivative as follows;
\begin{align}
    \frac{\partial \mathcal{L}_i}{\partial \theta_{ij}} = 
    \frac{-\beta\exp(\tilde{\delta}_{ij})\eta_{ij} + p_{ij}\sum_{k \in \mathcal{X}} \exp(\tilde{\delta}_{ik})\eta_{ik}}{\sum_{k \in \mathcal{X}} \exp(\tilde{\delta}_{ik})}
    \frac{\sum_{k \in \mathcal{X}} \exp(\tilde{\delta}_{ik})}{\sum_{k \in \mathcal{X}} \exp(\tilde{\delta}_{ik})\eta_{ik}}
    \frac{\sin(\theta_{ij} + m_1p_{ij})}{\tau}.
\end{align}
We will use $l$ to denote the index of a positive sample.
Furthermore, we use the fact that $\eta_{ih}$ of negative sample $h$ equals one.
We can further simplify the equation using this assumption as follows;
\begin{align}
    \frac{\partial \mathcal{L}_i}{\partial \theta_{ij}} = 
    \frac{-\beta\exp(\tilde{\delta}_{ij})\eta_{ij} + p_{ij}(\exp(\tilde{\delta}_{il})(\eta_{il}-1) + \sum_{k \in \mathcal{X}} \exp(\tilde{\delta}_{ik}))}{\sum_{k \in \mathcal{X}} \exp(\tilde{\delta}_{ik})}
    \frac{\sum_{k \in \mathcal{X}} \exp(\tilde{\delta}_{ik})}{\sum_{k \in \mathcal{X}} \exp(\tilde{\delta}_{ik})\eta_{ij}}
    \frac{\sin(\theta_{ij} + m_1p_{ij})}{\tau} \\
    = (p_{ij} + \frac{-\beta\exp(\tilde{\delta}_{ij})\eta_{ij} + p_{ij}\exp(\tilde{\delta}_{il})(\eta_{il}-1)}{\sum_{k \in \mathcal{X}} \exp(\tilde{\delta}_{ik})})
    \frac{\sum_{k \in \mathcal{X}} \exp(\tilde{\delta}_{ik})}{\sum_{k \in \mathcal{X}} \exp(\tilde{\delta}_{ik})\eta_{ij}}
    \frac{\sin(\theta_{ij} + m_1p_{ij})}{\tau} \\
    = (p_{ij} - \beta\tilde{q}_{ij}\eta_{ij} + p_{ij}\tilde{q}_{il}(\eta_{il}-1))
    \frac{\sum_{k \in \mathcal{X}} \exp(\tilde{\delta}_{ik})}{\sum_{k \in \mathcal{X}} \exp(\tilde{\delta}_{ik})\eta_{ij}}
    \frac{\sin(\theta_{ij} + m_1p_{ij})}{\tau} \\
    = (p_{ij} - \beta\tilde{q}_{ij}\eta_{ij} + p_{ij}\tilde{q}_{il}(\eta_{il}-1))
    \frac{\sum_{k \in \mathcal{X}} \exp(\tilde{\delta}_{ik})}{\exp(\tilde{\delta}_{il})(\eta_{il}-1) + \sum_{k \in \mathcal{X}} \exp(\tilde{\delta}_{ik})}
    \frac{\sin(\theta_{ij} + m_1p_{ij})}{\tau} \\
    = (p_{ij} - \beta\tilde{q}_{ij}\eta_{ij} + p_{ij}\tilde{q}_{il}(\eta_{il}-1))
    \frac{1}{1 - (1-\eta_{il})\tilde{q}_{il}}
    \frac{\sin(\theta_{ij} + m_1p_{ij})}{\tau}.
\end{align}

Since $\beta$ to one by the assumption, gradients for a positive sample $l$ can rewritten as follows;
\begin{align}
    \frac{\partial \mathcal{L}_i}{\partial \theta_{il}} = (p_{il} + \tilde{q}_{il}(p_{il}(\eta_{il}-1)- \beta\eta_{il}))
    \frac{1}{1 - (1-\eta_{il})\tilde{q}_{il}}
    \frac{\sin(\theta_{il} + m_1p_{il})}{\tau} \\
    = (p_{il} - \tilde{q}_{il})
    \frac{\sin(\theta_{il} + m_1p_{il})}{\tau}
    \frac{1}{1 - (1-\eta_{il})\tilde{q}_{il}}.
\end{align}

Similarly, by using $\eta_{ih}$ equals one, we can get the gradients for a negative sample $h$ as follows;
\begin{align}
    \frac{\partial \mathcal{L}_i}{\partial \theta_{ih}} = (p_{ih} - \beta\tilde{q}_{ih}\eta_{ih})
    \frac{1}{1 - (1-\eta_{il})\tilde{q}_{il}}
    \frac{\sin(\theta_{ih} + m_1p_{ih})}{\tau} \\
    = (p_{ih} - \tilde{q}_{ih})
    \frac{\sin(\theta_{ih} + m_1p_{ih})}{\tau}
    \frac{1}{1 - (1-\eta_{il})\tilde{q}_{il}}.
    \label{eq:subtractive}
\end{align}
That is, under the above assumptions, subtractive margin multiplies both positive and negative sample gradients by $1/(1 - (1-\eta_{il})\tilde{q}_{il})$.
If we reformulate the equation, we get the role of the subtractive margins:
\begin{align}
    \frac{\partial \mathcal{L}_i}{\partial \theta_{ij}} = (p_{ij} - \tilde{q}_{ij})
    \frac{\sin(\theta_{ij} + m_1p_{ij})}{\tau}
    \frac{1}{1 - (1-\eta_{il})\tilde{q}_{il}} \\
     = (p_{ij} - \tilde{q}_{ij})\frac{\sin(\theta_{ij})}{\sin(\theta_{ij})}
    \frac{\sin(\theta_{ij} + m_1p_{ij})}{\tau}
    \frac{1}{1 - (1-\eta_{il})\tilde{q}_{il}} \\
    = \frac{\partial \tilde{\mathcal{L}}_i}{\partial \theta_{ij}}
    \frac{\sin(\theta_{ij} + m_1p_{ij})}{\sin(\theta_{ij})}
    \frac{1}{1 - (1-\exp((\cos(\theta_{il} + m_1) - \cos(\theta_{il})-m_2)/\tau))\tilde{q}_{il}}.
\end{align}

\subsection{Lemma \ref{lemma:limit}}

As shown in Eq.~\ref{eq:subtractive}, $\partial \mathcal{L}_i/\partial \theta_{ij}$ can be expressed as follows:
\begin{align}
    \frac{\partial \mathcal{L}_i}{\partial \theta_{ij}}
    = (p_{ij} - \tilde{q}_{ij})
    \frac{\sin(\theta_{ij} + m_1p_{ij})}{\tau}
    \frac{1}{1 - (1-\exp((\cos(\theta_{il} + m_1) - \cos(\theta_{il})-m_2)/\tau))\tilde{q}_{il}}.
\end{align}
Therefore, we can get the following equation for a positive sample $l$:
\begin{align}
    \lim_{m_2\to\infty} \frac{\partial \mathcal{L}_i}{\partial \theta_{il}} 
    = \lim_{m_2\to\infty} (p_{il} - \tilde{q}_{il})
    \frac{\sin(\theta_{il} + m_1p_{il})}{\tau (1 - (1-\exp((\cos(\theta_{il} + m_1) - \cos(\theta_{il})-m_2)/\tau))\tilde{q}_{il})} \\
    = (p_{il} - \tilde{q}_{il})
    \frac{\sin(\theta_{il} + m_1p_{il})}{(1- \tilde{q}_{il})\tau} = 
    \frac{\sin(\theta_{il} + m_1)}{\tau}.
\end{align}



\begin{table*}[ht]
    \begin{minipage}{0.48\textwidth}
        \centering
        \caption{Margin values for four datasets (Tab.~\ref{tab:main_frozen}).}
        \begin{tabular}{cccc}
            \toprule
            Method & Dataset & \textit{m1} & \textit{m2} \\
            \midrule
            \textit{MoCov3} & CIFAR10        & 0.1 & 0.4 \\
                          & CIFAR100       & 0.5 & 0.7 \\
                          & STL10          & 0.4 & 0.6 \\
                          & TinyImageNet  & 0.1 & 0.4 \\
            \midrule
            \textit{SimCLR} & CIFAR10      & 0.5 & 0.4 \\
                          & CIFAR100       & 0.6 & 0.6 \\
                          & STL10          & 0.0 & 0.2 \\
                          & TinyImageNet  & 0.0 & 0.8 \\
            \bottomrule
            \label{tab:hparam_margins}
        \end{tabular}
    \end{minipage}
    \hfill
    \begin{minipage}{0.48\textwidth}
        \centering
        \caption{Scaling factors of $pos.$ only models.}
        \begin{tabular}{ccc}
            \toprule
            Method & Dataset & \textit{pos.} ($s$) \\
            \midrule
            \textit{MoCov3} & CIFAR10        & 20 \\
                          & CIFAR100       & 20 \\
                          & STL10          & 40 \\
                          & TinyImageNet  & 20  \\
            \midrule
            \textit{SimCLR} & CIFAR10      & 1.125  \\
                          & CIFAR100       & 1.0625  \\
                          & STL10          & 1.0625  \\
                          & TinyImageNet  & 1.0625  \\
            \bottomrule
            \label{tab:hparam_pos}
        \end{tabular}
    \end{minipage}
    \begin{minipage}{0.48\textwidth}
        \centering
        \caption{Scaling factors and curvature factors.}
        \begin{tabular}{cccc}
            \toprule
            Method & Dataset & \textit{pos.} ($s$) & \textit{curv.} ($c$) \\
            \midrule
            \textit{MoCov3} & CIFAR10        & 20 & 0.7 \\
                          & CIFAR100       & 40 & 1 \\
                          & STL10          & 40 & 2.5 \\
                          & TinyImageNet  & 40 & 2.5 \\
            \midrule
            \textit{SimCLR} & CIFAR10      & 2.5 & 0.7 \\
                          & CIFAR100       & 2 & 0.7 \\
                          & STL10          & 1.25 & 0.7 \\
                          & TinyImageNet  & 1.125 & 1.0 \\
            \midrule
            \textit{BYOL} & CIFAR10 & - & 1.0 \\
                          & CIFAR100 & - & 2.5 \\
                          & STL10 & - & 5.0 \\
                          & TinyImageNet & - & 2.5 \\
            \bottomrule
            \label{tab:hparam_pos_curv}
        \end{tabular}
    \end{minipage}
    \hfill
    \begin{minipage}{0.48\textwidth}
        \centering
        \caption{Angular margin $m_1$ to calculate the ratio (Tabs.~\ref{tab:main_frozen} and \ref{tab:main-transfer})}
        \begin{tabular}{ccc}
            \toprule
            Method & Dataset & $m_1$ \\
            \midrule
            \textit{MoCov3} & CIFAR10        & 1.6 \\
                          & CIFAR100       & 1.6 \\
                          & STL10          & 0.4 \\
                          & TinyImageNet  & 1.6 \\
            \midrule
            \textit{SimCLR} & CIFAR10      & 0.2 \\
                          & CIFAR100       & 0.2 \\
                          & STL10          & 0.4 \\
                          & TinyImageNet  & 0.2 \\
            \bottomrule
            \label{tab:hparam_pos_curv}
        \end{tabular}
    \end{minipage}
\end{table*}

\clearpage
\section{Visualization}

\begin{figure*}[]
    \centering
    \includegraphics[width=\linewidth]{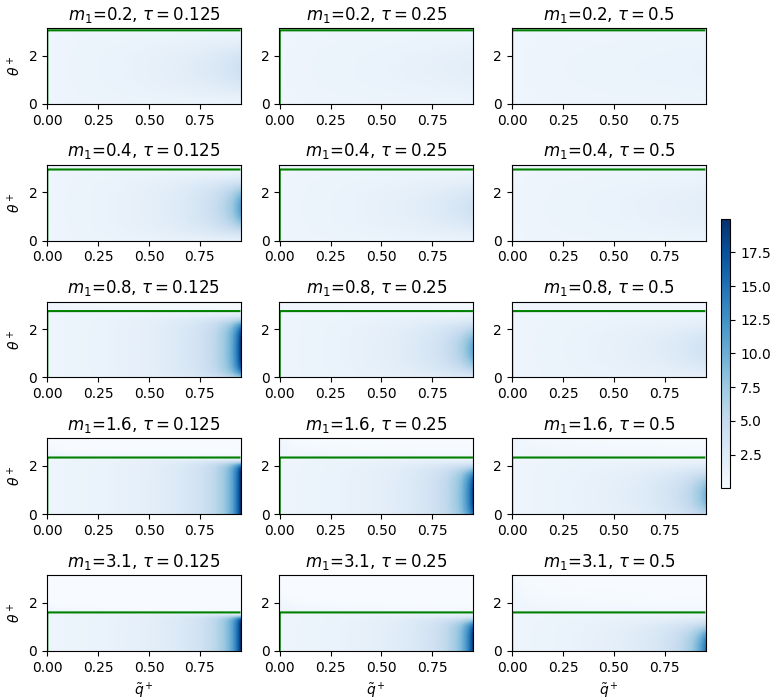}
    \vspace{-0.5cm}
    \caption{Visualization of $(p_{ij} - \beta q_{ij})/ (p_{ij} - \beta \tilde{q}_{ij})$. The green lines indicate where the ratio is one. $m_1$ and $\tau$ are the angular margin and temperature, respectively (Eq.~\ref{eq:delta}).}
    \label{fig:appendix-the-ratio}
\end{figure*}

Fig.~\ref{fig:appendix-the-ratio} visualizes the weight distributions of the ratios $(p_{ij} - \beta q_{ij})/ (p_{ij} - \beta \tilde{q}_{ij})$.
Angular margin $m_1$ and temperature $\tau$ are from Eq.~\ref{eq:delta}.
The figure shows that as $m_1$ increases, the center of emphasis in $\theta^+$ lowers and the emphasizing area broadens.

\section{Implememtation Details}
\label{sec:app_imp_details}

\subsection{Experimental settings for CIFAR-10, CIFAR-100, STL-10, and TinyImageNet}
Our experimental settings for these dataset basically follows that of ReSSL.
We pretrained each model on a source dataset for 200 epochs and then fine-tuned it on a target dataset for 100 epochs.
We set the batch size to 256 and $\tau$ to 0.25 (for MoCov3 and SimCLR).
The latent feature dimension was set to 128, while the hidden dimension of linear layers (including projection heads and predictors) was set to 2048.
We used SGD with a momentum of 0.9 for both pretraining and evaluation.
The learning rate was set to 0.06 for pretraining the backbone model and 1 for fine-tuning the linear layer.
During evaluation, we employed the cosine annealing learning rate scheduler.

We used two augmentation policies: strong and weak augmentation.
The strong augmentation consists of random resize cropping, horizontal flipping, color jittering, random gray scaling, and Gaussian blurring.
In contrast, the weak augmentation only included random horizontal and random cropping.
For contrastive learning-based SSL methods with a teacher model (MoCov3, BYOL), we used weak data augmentation to the teacher model and strong data augmentation to the student model.
For a method without a teacher model (SimCLR), however, we use these two different augmentation policies to generate two different representations of the same identity.

For projection heads and predictors of MoCov3, SimCLR, and BYOL, we used batch normalization before the ReLU activation.

During hyperparameter tuning for comparison with baselines, we only adjusted margins ($m_1$ and $m_2$), $s$ (Eq.\ref{eq:pos-emp-overall}), and $c$ (Eq.\ref{eq:curve-control}), while keeping other hyperparameters fixed, such as the learning rate.
The hyperparameters used in Sec.~\ref{ssec:results-comp-baselines} are specified in Tabs.~\ref{tab:hparam_margins}, \ref{tab:hparam_pos_curv} and \ref{tab:hparam_pos}.

\subsection{Experimental settings for ImageNet}
We used the official implementations of both MoCov3 and SimCLR.
Due to our computational budget, we pretrained the networks for 100 epochs.
Following the experimental settings of MoCov3 and SimCLR, we fine-tuned the last linear layer for 90 epochs for evaluation.
We only modified the gradient scales using Alg.~\ref{alg:pos_curv}, without making any other changes.
When training SimCLR, we used a two-linear layered projection module.
This choice was made because only the official performance of the two-linear-layered version, pretrained for 100 epochs, was available.
For both MoCo and SimCLR, we use a batch size of 4,096.

The hyperparameters ($s$ and $c$), used in Tab.~\ref{tab:imagenet} are as follows: for MoCov3, we set $s$ to 10 and $c$ to 1.5, while for SimCLR, $s$ was set to 2 and $c$ to 1.

\section{Additional Experiments}

We ran additional experiments on MoCov3 for much longer epochs (1,000).
Tabs.~\ref{tab:linear-prob-1000} and \ref{tab:transfer-1000} show that emphasizing positive samples improves performance not only in the short run but even in the long run.

\begin{table*}[ht]
\centering
\caption{Linear Probing of MoCov3.}
\begin{tabular}{cccccc}
    \toprule
    $s$ & epochs & CIFAR10 & CIFAR100 & STL10 & TinyImageNet \\
     \midrule
    1 & 1000 & 91.142 $\pm$ 0.218 & 65.010 $\pm$ 0.215 & 90.280 $\pm$ 0.113 & 45.682 $\pm$ 0.151\\
    10 & 1000 & 92.386 $\pm$ 0.100 & 69.136 $\pm$ 0.421 & 90.633 $\pm$ 0.244 & 50.478 $\pm$ 0.237\\
    20 & 1000 & 92.624 $\pm$ 0.161 & \textbf{69.466 $\pm$ 0.183} & 90.860 $\pm$ 0.122 & \textbf{51.154 $\pm$ 0.184} \\
    40 & 1000 & \textbf{92.648 $\pm$ 0.171} & 69.408 $\pm$ 0.342 & \textbf{91.098 $\pm$ 0.209} & 51.138 $\pm$ 0.362 \\
    \bottomrule
    \label{tab:linear-prob-1000}
\end{tabular}
\end{table*}

\begin{table*}[ht]
\centering
\caption{Transfer Learning of MoCov3.}
\begin{tabular}{cccccc}
    \toprule
     & & CIFAR100 & CIFAR10 & TinyImageNet & STL10 \\
    $s$ & epochs & $\xrightarrow{}$ CIFAR10 & $\xrightarrow{}$ CIFAR100 & $\xrightarrow{}$ STL10 & $\xrightarrow{}$ TinyImageNet \\
    \midrule
    1 & 1000 & 75.298 $\pm$ 0.310 & 40.696 $\pm$ 0.327 & 71.088 $\pm$ 0.361 & 30.358 $\pm$ 0.304 \\
    10 & 1000 & 79.858 $\pm$ 0.255 & 53.028 $\pm$ 0.569 & 77.533 $\pm$ 0.101 & 38.702 $\pm$ 0.420 \\
    20 & 1000 & \textbf{80.144 $\pm$ 0.105} & 54.515 $\pm$ 0.235 & 78.623 $\pm$ 0.193 & 40.464 $\pm$ 0.341 \\
    40 & 1000 & 79.800 $\pm$ 0.162 & \textbf{54.612 $\pm$ 0.317} & \textbf{78.810 $\pm$ 0.198} & \textbf{40.600 $\pm$ 0.360} \\
    
    \bottomrule
    \label{tab:transfer-1000}
\end{tabular}
\end{table*}

\section{Pseudo code}
\label{sec:app_pseudo_code}

In this section, we provide pseudo codes for reproduction.
Although contrastive learning-based baselines do not explicitly utilize angles, as mentioned in the main paper, computing logits using cosine similarity can be interpreted as obtaining logits using the angles that result from taking the arccos of the cosine similarity values (Alg.~\ref{alg:org}).
We used Alg.~\ref{alg:m1m2} to use margins in contrastive learning methods.
To scale the gradients, it can be accomplished by replacing only the process of converting angles to logits from Alg.~\ref{alg:org} to Algs.~\ref{alg:pos_curv}, \ref{alg:ratio}, \ref{alg:effect3_typeI}, and \ref{alg:effect3_typeII}.
As the algorithms demonstrate, our modification does not affect the logit values but solely modifies the gradients.
To adjust the curvature of positive gradient scales in BYOL, it suffices to set $p$ as an all-ones vector since only positive samples are used.

\begin{algorithm}
    \caption{Logits as a function of angles: pseudo code of general contrastive learning (Eq.~\ref{eq:delta-ori})}
    \label{alg:org}
    \begin{algorithmic}
        \State \textbf{Inputs:}
        \Statex \hspace*{\algorithmicindent}\parbox[t]{.8\linewidth}{\raggedright $\theta$ \ \ \ \ \ \ \ \  Angles}
        \Statex \hspace*{\algorithmicindent}\parbox[t]{.8\linewidth}{\raggedright $\tau$ \ \ \ \ \ \ \ \ Temperature}
        \State $logits \gets \cos(\theta) / \tau$
        \State \textbf{return} $logits$
    \end{algorithmic}
\end{algorithm}

\begin{algorithm}
    \caption{Logits with margins ($m_1$ and $m_2$) (Eq.~\ref{eq:delta})}
    \label{alg:m1m2}
    \begin{algorithmic}
        \State \textbf{Inputs:}
        \Statex \hspace*{\algorithmicindent}\parbox[t]{.8\linewidth}{\raggedright $\theta$ \quad \quad Angles}
        \Statex \hspace*{\algorithmicindent}\parbox[t]{.8\linewidth}{\raggedright $\tau$ \quad \quad Temperature}
        \Statex \hspace*{\algorithmicindent}\parbox[t]{.8\linewidth}{\raggedright $p$ \quad \quad Target probabilities}
        \Statex \hspace*{\algorithmicindent}\parbox[t]{.8\linewidth}{\raggedright $m_1$ \quad \  Angular margin}
        \Statex \hspace*{\algorithmicindent}\parbox[t]{.8\linewidth}{\raggedright $m_2$ \quad \  Subtractive margin}
        \State $logits \gets (\cos(\theta + p\times m_1) - p\times m_2)/ \tau$
        \State \textbf{return} $logits$
    \end{algorithmic}
\end{algorithm}

\begin{algorithm}
    \caption{Emphasizing positive samples and controlling the curvature of positive gradient scales (Fig.~\ref{fig:sc-curve-acc}, Tabs.~\ref{tab:imagenet} and \ref{tab:main_frozen}).}
    \label{alg:pos_curv}
    \begin{algorithmic}
        \State \textbf{Inputs:}
        \Statex \hspace*{\algorithmicindent}\parbox[t]{.8\linewidth}{\raggedright $\theta$ \quad \quad Angles}
        \Statex \hspace*{\algorithmicindent}\parbox[t]{.8\linewidth}{\raggedright $\tau$ \quad \quad Temperature}
        \Statex \hspace*{\algorithmicindent}\parbox[t]{.8\linewidth}{\raggedright $p$ \quad \quad Target probabilities}
        \Statex \hspace*{\algorithmicindent}\parbox[t]{.8\linewidth}{\raggedright $s$ \quad \quad  Scaling factor}
        \Statex \hspace*{\algorithmicindent}\parbox[t]{.8\linewidth}{\raggedright $c$ \quad \quad  Curvature factor}
        \State $logits \gets \cos(\theta) / \tau$
        \State $scales \gets stop\_gradient(s \times (1 - \frac{\theta}{\pi} ^c)^{1/c})$
        \State $logits \gets (1-p) \times logits + p \times ((1 - scales)\times stop\_gradient(logits) + scales\times logits)$
        \State \textbf{return} $logits$
    \end{algorithmic}
\end{algorithm}

\begin{algorithm}
    \caption{Scaling gradients by the ratio (Fig.~\ref{fig:third-ratio}).}
    \label{alg:ratio}
    \begin{algorithmic}
        \State \textbf{Inputs:}
        \Statex \hspace*{\algorithmicindent}\parbox[t]{.8\linewidth}{\raggedright $\theta$ \quad \quad Angles}
        \Statex \hspace*{\algorithmicindent}\parbox[t]{.8\linewidth}{\raggedright $\tau$ \quad \quad Temperature}
        \Statex \hspace*{\algorithmicindent}\parbox[t]{.8\linewidth}{\raggedright $p$ \quad \quad Target probabilities}
        \Statex \hspace*{\algorithmicindent}\parbox[t]{.8\linewidth}{\raggedright $m_1$ \quad \  Angular margin}
        
        \State $logits \gets \cos(\theta) / \tau$
        \State $new\_logits \gets \cos(\theta + p\times m_1)/ \tau$
        \State $ratio \gets sum(\exp(logits), dim=-1) / sum(\exp(new\_logits), dim=-1)$
        \State $scales \gets stop\_gradient(ratio)$
        \State $logits \gets (1-p) \times logits + p \times ((1 - scales)\times stop\_gradient(logits) + scales\times logits)$
        \State \textbf{return} $logits$
    \end{algorithmic}
\end{algorithm}

\begin{algorithm}
    \caption{Attenuating the diminishing gradients (Type I) (Eq.~\ref{eq:type1})}
    \label{alg:effect3_typeI}
    \begin{algorithmic}
        \State \textbf{Inputs:}
        \Statex \hspace*{\algorithmicindent}\parbox[t]{.8\linewidth}{\raggedright $\theta$ \quad \quad Angles}
        \Statex \hspace*{\algorithmicindent}\parbox[t]{.8\linewidth}{\raggedright $\tau$ \quad \quad Temperature}
        \Statex \hspace*{\algorithmicindent}\parbox[t]{.8\linewidth}{\raggedright $p$ \quad \quad Target probabilities}
        \Statex \hspace*{\algorithmicindent}\parbox[t]{.8\linewidth}{\raggedright $\alpha$ \quad \quad  Attenuation factor}
        \State $logits \gets \cos(\theta) / \tau$
        \State $q \gets softmax(logits)$
        \State $scales \gets sum(p / stop\_gradient(p - \alpha \times q),\ dim=-1)$
        \State $logits \gets (1-scales) \times stop\_gradient(logits) + scales \times logits$
        \State \textbf{return} $logits$
    \end{algorithmic}
\end{algorithm}

\begin{algorithm}
    \caption{Attenuating the diminishing gradients of positive samples (Type II) (Eq.~\ref{eq:type1})}
    \label{alg:effect3_typeII}
    \begin{algorithmic}
        \State \textbf{Inputs:}
        \Statex \hspace*{\algorithmicindent}\parbox[t]{.8\linewidth}{\raggedright $\theta$ \quad \quad Angles}
        \Statex \hspace*{\algorithmicindent}\parbox[t]{.8\linewidth}{\raggedright $\tau$ \quad \quad Temperature}
        \Statex \hspace*{\algorithmicindent}\parbox[t]{.8\linewidth}{\raggedright $p$ \quad \quad Target probabilities}
        \Statex \hspace*{\algorithmicindent}\parbox[t]{.8\linewidth}{\raggedright $\alpha$ \quad \quad  Attenuation factor}
        \State $logits \gets \cos(\theta) / \tau$
        \State $q \gets softmax(logits)$
        \State $scales \gets 1 / stop\_gradient(p - \alpha \times q)$
        \State $logits \gets (1-p) \times logits + p \times ((1 - scales)\times stop\_gradient(logits) + scales\times logits)$
        \State \textbf{return} $logits$
    \end{algorithmic}
\end{algorithm}

\end{document}